\documentclass[final]{cvpr} 

\usepackage{times}
\usepackage{epsfig}
\usepackage{graphicx}
\usepackage{amsmath}
\usepackage{amssymb}
\usepackage{booktabs}
\usepackage[numbers]{natbib}

\usepackage[pagebackref=true,breaklinks=true,colorlinks,bookmarks=false]{hyperref}

\def\cvprPaperID{****} 
\def\confYear{CVPR 2021}

\begin{document}

\title{Modular Framework for Visuomotor Language Grounding}

\author{Kolby Nottingham \and Litian Liang \and Daeyun Shin \and Charless C. Fowlkes \and Roy Fox \and Sameer Singh\\
University of California Irvine\\
{\tt\small \{knotting,litianl1,daeyuns,fowlkes,royf,sameer\}@uci.edu}
}

\maketitle

\begin{abstract}
Natural language instruction following tasks serve as a valuable test-bed for grounded language and robotics research. However, data collection for these tasks is expensive and end-to-end approaches suffer from data inefficiency. We propose the structuring of language, acting, and visual tasks into separate modules that can be trained independently. Using a Language, Action, and Vision (LAV) framework removes the dependence of action and vision modules on instruction following datasets, making them more efficient to train. We also present a preliminary evaluation of LAV on the ALFRED task for visual and interactive instruction following.
\end{abstract}

\section{Introduction}

Many state of the art natural language systems are conditioned solely on language input \cite{brown2020language,devlin2018bert,raffel2019exploring}. However advanced language understanding requires that language is grounded in vision and interaction \cite{bender2020climbing,bisk2020experience}. Interactive and visual instruction following tasks provide a test-bed for developing methods that ground language in vision and actions. These types of tasks are also interesting from a robotics perspective. Ideally robots that interact with humans in the real world will support a natural language interface. Thus interactive and visual natural language instruction following tasks also work towards accomplishing this goal.

Typical approaches to instruction following tasks perform end-to-end learning with a deep neural network \cite{anderson2018vision,shridhar2020alfred}. However, gathering expert demonstrations paired with natural language instructions is costly and datasets are typically small. End-to-end baselines have performed poorly on interactive instruction following tasks, and many approaches to improve on baselines incorporate some modularization. \citet{corona2020modularity} train separate modules for each type of high level task to be completed (e.g., go to, pick up), and \citet{singh2020moca} train a perception module separate from their action module. However, all modules in both of these methods are still dependent on the instruction following dataset. Recently, \citet{saha2021modular} introduced a method called Modular Vision and Language (MoViLan) that is most similar to our method. MoViLan also trains modules independently to work together at test time, but it is evaluated on a simpler version of the ALFRED task.

Simple end-to-end approaches for solving instruction following tasks ignore the fact that much can be learned about actions and vision independent of instruction datasets. We make this separation explicit and propose a Language, Action, and Vision (LAV) framework that can train each module independently, thus removing the action and vision modules' dependence on any instruction following dataset. Our evaluation of LAV on the ALFRED task indicates that it is able to significantly outperform end-to-end baselines.

\begin{figure}
    \centering
    \includegraphics[width=\linewidth]{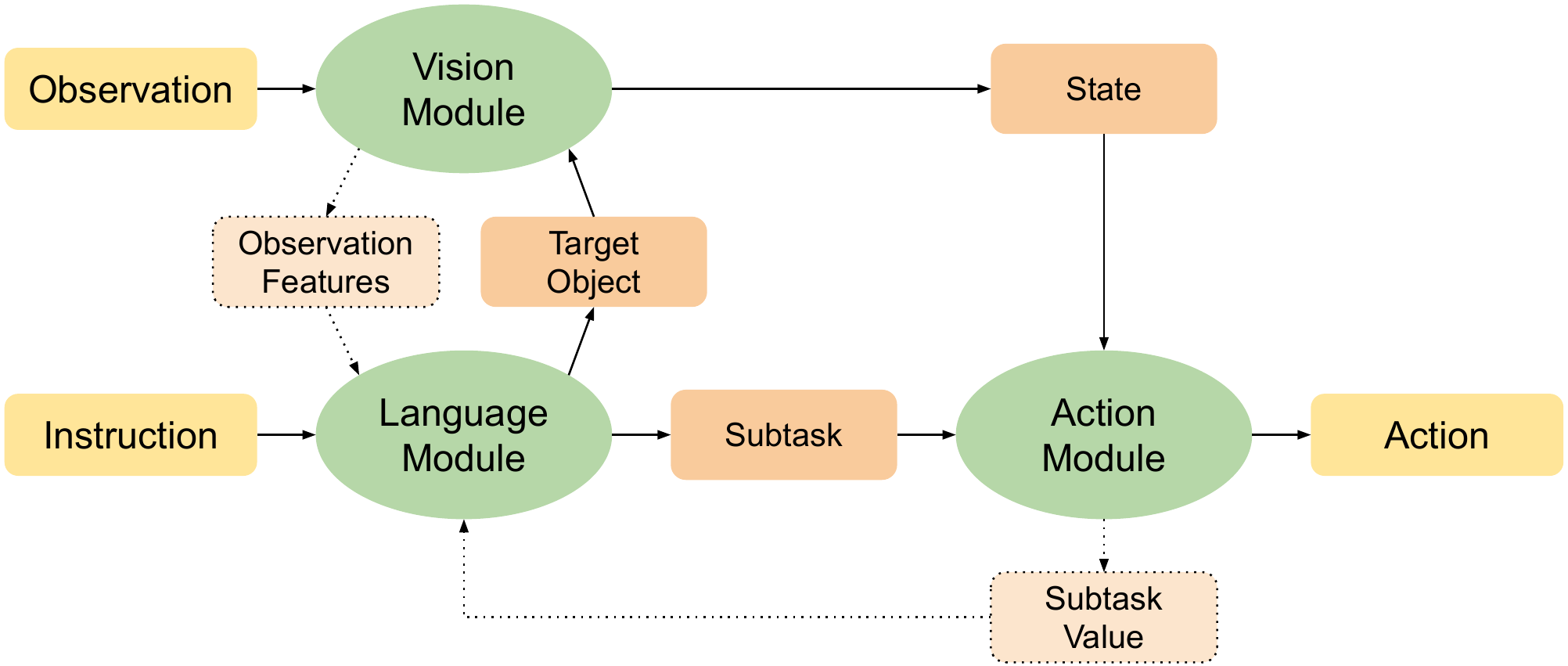}
    \caption{The Language Action Vision (LAV) framework. Modules are trained independently to minimize dependence on natural language instructions and expert demonstrations. Items outlined with a dotted line are suggested improvements.}
    \label{fig:lav}
\end{figure}


\section{LAV Framework}

Key to our LAV framework (see Figure \ref{fig:lav}) is that the action and vision modules can be trained independently of natural language instructions and expert trajectories. This usually means defining a set of possible action subtasks $g$ and target objects $o$. Given the natural language instruction $x_{lang}$, the main task of the \textbf{Language Module} $L$ is to identify the subtask and target object for the action and vision modules $L: x_{lang} \to o, g$.

With a set of possible objects, the LAV \textbf{Vision Module} $V$ can be trained to extract state features $s$ given visual observations $x_{obs}$ and a target object $o$, independent of the instruction following dataset $V: x_{obs}, o \to s$. The vision module benefits from this independence because computer vision datasets are typically cheaper to collect and more plentiful than instruction following datasets. Additionally, many instruction following tasks run in simulation making vision datasets especially simple to collect and label.

Finally the LAV \textbf{Action Module} $A$ can learn to complete subtasks with arbitrary target objects in a multi-task learning setting. Given a subtask $g$, target object $o$, and state featues $s$, it learns to predict actions $a$ to complete the subtask $A: g, o, s \to a$. Multi-task robot learning is a popular research area and many approaches exist for solving this problem. For example, if the tasks are running in a simulator, the action module can be trained via multi-task reinforcement learning.

Note that the output from each independent module can also augment the input to other modules to improve test time performance as illustrated by the dotted items in Figure \ref{fig:lav}. For example, the language module can be informed by what objects are visible in the scene ($L: x_{lang}, s' \to o, g$ where $V': x_{obs} \to s'$). Additionally, the language module can use the value of the current state under a specific subtask to determine the output with the highest chance of success ($L: x_{lang}, v \to  o, g$ where $A': s' \to v$).

\section{Evaluation}

To test our LAV framework, we develop a prototype implementation to evaluate on the ALFRED task \cite{shridhar2020alfred}.

\subsection{Implementation}

The LAV language module for our implementation is finetuned from a T5 language model \cite{raffel2019exploring}. We train the model to generate sequences of (subtask, target object) pairs based on natural language goal instructions. The subtasks we use consist of the seven high level actions defined by ALFRED (pick up, place, toggle, clean, cool, heat, and slice), and the objects consist of all of the object types used as target objects in the training set.

We train three vision networks that make up our LAV vision module. Given an RGB observation, these networks output an object type segmentation, a depth map, and an obstacle indicator. All three models were trained via supervised learning from datasets collected from the AI2THOR simulator. Depth maps are used in with segmentations to estimate the relative position to a target object.

Using the target object's position estimated by the vision module, the LAV action module first navigates toward the target object and then executes low level actions corresponding to the current subtask predicted by the language module. Navigation is a simple depth first search around obstacles toward the target object's estimated position.

For example, after picking up a dirty bowl the language module predicts the ``clean'' subtask and the ``sink'' target object. The vision module identifies the sink and provides an estimated position. The action module navigates toward the sink. Once the estimated position is in range, the action module performs the actions ``place in sink'', ``toggle sink on'', ``toggle sink off'', and ``pick up bowl''.

\begin{table}[t]
\small
\setlength{\tabcolsep}{2.75pt}
\center
\begin{tabular}{lcccccccc}
\toprule
& \multicolumn{4}{c}{\bf Test Data Seen}  & \multicolumn{4}{c}{\bf Test Data Unseen} \\
\cmidrule(lr){2-5}
\cmidrule(lr){6-9}
                               & SR   & PWSR & GC   & PWGC & SR  & PWSR & GC   & PWGC \\
\midrule
Baseline                       & 4.0  & 2.0  & 9.4  & 6.3  & 0.4 & 0.1  & 7.0  & 4.3  \\
LWIT \cite{alfred_leaderboard} & 30.9 & 25.9 & 40.5 & 36.8 & 9.4 & 5.6  & 20.9 & 16.3 \\
\bf LAV                        & 13.4 & 6.3  & 23.2 & 13.2 & 6.3 & 3.1  & 17.3 & 10.5 \\
\bottomrule
\end{tabular}
\caption{Percent success rate (SR), path weighted success rate (PWSR), goal condition success rate (GC), and path weighted goal condition success rate (PWGC) from ALFRED's public leaderboard. Seen test data indicates scenes included in the training set but with novel tasks while unseen data used novel scenes.}
\label{table:leaderboard}
\end{table}

\begin{table}[t]
\small
\center
\begin{tabular}{lcccc}
\toprule
                           & SR   & PWSR & GC   & PWGC \\
\midrule
L \& V Oracles             & 25.2 & 8.5  & 32.5 & 11.3 \\
V Oracle                   & 18.4 & 6.3  & 26.3 & 8.6  \\
L Oracle                   & 15.4 & 7.3  & 24.8 & 15.3 \\
\bf LAV                    & 12.7 & 5.9  & 23.4 & 13.7 \\
\bottomrule
\end{tabular}
\caption{Metrics comparing versions of LAV on ALFRED's validation data (seen). Oracles replace a module with ground truth.}
\label{table:oracles}
\end{table}

\subsection{Results}

We compare our LAV framework to the ALFRED baseline and the current state of the art (LWIT \cite{alfred_leaderboard}) in Table \ref{table:leaderboard}. We also provide results when replacing our language and vision modules with ground truth oracles in Table \ref{table:oracles}. The LAV framework significantly outperforms the baseline across all metrics. While it doesn't outperform LWIT in its current form, LAV suffers less from the transfer to novel test scenes than LWIT does. Also note that LAV only achieves 25.2\% SR while using language and vision oracles. This indicates that the current point navigation search is a major weak point of our implementation. We plan to replace the current action module with a reinforcement learning agent in the future which will further improve performance.

\section{Conclusion}

The LAV framework demonstrates the advantage of training vision and action modules independent of instruction datasets. Doing so allows those modules to train on much cheaper and more abundant data. The language module is able to predict subtasks and target objects from instruction data without needing to learn vision and low level actions as well. In the future we plan to apply LAV to other tasks such as iGibson \cite{xia2020interactive} and AI2-THOR Rearrangement \cite{RoomR} and improve upon the current LAV implementation.

{\small
\bibliographystyle{ieee_fullname_nat}
\bibliography{egbib}

\begin{thebibliography}{10}\itemsep=-1pt

\bibitem[alf()]{alfred_leaderboard}
Alfred leaderboard.
\newblock \url{https://leaderboard.allenai.org/alfred/submissions/public}.
\newblock Accessed: 2021-05-14.

\bibitem[Anderson et~al.(2018)]{anderson2018vision}
Peter Anderson, Qi Wu, Damien Teney, Jake Bruce, Mark Johnson, Niko
  S{\"u}nderhauf, Ian Reid, Stephen Gould, and Anton Van Den~Hengel.
\newblock Vision-and-language navigation: Interpreting visually-grounded
  navigation instructions in real environments.
\newblock In {\em Proceedings of the IEEE Conference on Computer Vision and
  Pattern Recognition}, pages 3674--3683, 2018.

\bibitem[Bender and Koller(2020)]{bender2020climbing}
Emily~M Bender and Alexander Koller.
\newblock Climbing towards nlu: On meaning, form, and understanding in the age
  of data.
\newblock In {\em Proceedings of the 58th Annual Meeting of the Association for
  Computational Linguistics}, pages 5185--5198, 2020.

\bibitem[Bisk et~al.(2020)]{bisk2020experience}
Yonatan Bisk, Ari Holtzman, Jesse Thomason, Jacob Andreas, Yoshua Bengio, Joyce
  Chai, Mirella Lapata, Angeliki Lazaridou, Jonathan May, Aleksandr Nisnevich,
  et~al.
\newblock Experience grounds language.
\newblock {\em arXiv preprint arXiv:2004.10151}, 2020.

\bibitem[Brown et~al.(2020)]{brown2020language}
Tom~B Brown, Benjamin Mann, Nick Ryder, Melanie Subbiah, Jared Kaplan, Prafulla
  Dhariwal, Arvind Neelakantan, Pranav Shyam, Girish Sastry, Amanda Askell,
  et~al.
\newblock Language models are few-shot learners.
\newblock {\em arXiv preprint arXiv:2005.14165}, 2020.

\bibitem[Corona et~al.(2020)]{corona2020modularity}
Rodolfo Corona, Daniel Fried, Coline Devin, Dan Klein, and Trevor Darrell.
\newblock Modularity improves out-of-domain instruction following.
\newblock {\em arXiv preprint arXiv:2010.12764}, 2020.

\bibitem[Devlin et~al.(2018)]{devlin2018bert}
Jacob Devlin, Ming-Wei Chang, Kenton Lee, and Kristina Toutanova.
\newblock Bert: Pre-training of deep bidirectional transformers for language
  understanding.
\newblock {\em arXiv preprint arXiv:1810.04805}, 2018.

\bibitem[Raffel et~al.(2019)]{raffel2019exploring}
Colin Raffel, Noam Shazeer, Adam Roberts, Katherine Lee, Sharan Narang, Michael
  Matena, Yanqi Zhou, Wei Li, and Peter~J Liu.
\newblock Exploring the limits of transfer learning with a unified text-to-text
  transformer.
\newblock {\em arXiv preprint arXiv:1910.10683}, 2019.

\bibitem[Saha et~al.(2021)]{saha2021modular}
Homagni Saha, Fateme Fotouhif, Qisai Liu, and Soumik Sarkar.
\newblock A modular vision language navigation and manipulation framework for
  long horizon compositional tasks in indoor environment.
\newblock {\em arXiv preprint arXiv:2101.07891}, 2021.

\bibitem[Shridhar et~al.(2020)]{shridhar2020alfred}
Mohit Shridhar, Jesse Thomason, Daniel Gordon, Yonatan Bisk, Winson Han,
  Roozbeh Mottaghi, Luke Zettlemoyer, and Dieter Fox.
\newblock Alfred: A benchmark for interpreting grounded instructions for
  everyday tasks.
\newblock In {\em Proceedings of the IEEE/CVF conference on computer vision and
  pattern recognition}, pages 10740--10749, 2020.

\bibitem[Singh et~al.(2020)]{singh2020moca}
Kunal~Pratap Singh, Suvaansh Bhambri, Byeonghwi Kim, Roozbeh Mottaghi, and
  Jonghyun Choi.
\newblock Moca: A modular object-centric approach for interactive instruction
  following.
\newblock {\em arXiv preprint arXiv:2012.03208}, 2020.

\bibitem[Weihs et~al.(2021)]{RoomR}
Luca Weihs, Matt Deitke, Aniruddha Kembhavi, and Roozbeh Mottaghi.
\newblock Visual room rearrangement.
\newblock In {\em IEEE/CVF Conference on Computer Vision and Pattern
  Recognition (CVPR)}, June 2021.

\bibitem[Xia et~al.(2020)]{xia2020interactive}
Fei Xia, William~B Shen, Chengshu Li, Priya Kasimbeg, Micael~Edmond Tchapmi,
  Alexander Toshev, Roberto Mart{\'\i}n-Mart{\'\i}n, and Silvio Savarese.
\newblock Interactive gibson benchmark: A benchmark for interactive navigation
  in cluttered environments.
\newblock {\em IEEE Robotics and Automation Letters}, 5(2):713--720, 2020.

\end{thebibliography}
}

\end{document}